\PassOptionsToPackage{table,xcdraw}{xcolor}
\documentclass[manuscript,screen]{acmart}
\AtBeginDocument{%
  \providecommand\BibTeX{{%
    \normalfont B\kern-0.5em{\scshape i\kern-0.25em b}\kern-0.8em\TeX}}}

\begin{document}

\title{Explainability in Deep Reinforcement Learning, a Review into Current Methods and Applications}

\author{Thomas Hickling}
\orcid{0000-0001-5317-5021}
\affiliation{%
  \institution{Department of Electrical Engineering, City University of London}
  \streetaddress{Northampton Square}
  \city{London}
  \postcode{EC1V 0HB}
  \country{UK}
}
\email{tom.hickling@city.ac.uk}
\author{Abdelhafid Zenati}
\affiliation{%
  \institution{Department of Electrical Engineering, City University of London}
  \city{London}
  \country{UK}
}
\email{abdelhafid.zenati@city.ac.uk}
\author{Nabil Aouf}
\affiliation{%
  \institution{Department of Electrical Engineering, City University of London}
  \city{London}
  \country{UK}
}
\email{nabil.aouf@city.ac.uk}

\author{Phillippa Spencer}
\affiliation{%
  \institution{Defence, Science and Technology Laboratory (Dstl)}
  \city{Salisbury}
  \country{UK}
}

\renewcommand{\shortauthors}{Hickling et al.}

{\color{black}\begin{abstract}
    The use of Deep Reinforcement Learning (DRL) schemes has increased dramatically since their first introduction in 2015. Though uses in many different applications are being found, they still have a problem with the lack of interpretability. This has bread a lack of understanding and trust in the use of DRL solutions from researchers and the general public. To solve this problem, the field of Explainable Artificial Intelligence (XAI) has emerged. This entails a variety of different methods that look to open the DRL black boxes, ranging from the use of interpretable symbolic Decision Trees (DT) to numerical methods like Shapley Values. This review looks at which methods are being used and for which applications. This is done to identify which models are the best suited to each application or if a method is being underutilised. 
\end{abstract}



\maketitle

\section{Introduction}
The use of Machine Learning (ML) to optimise solutions in commercial and public projects has increased rapidly over the past decade. Tasks such as weather simulation, medical diagnosis, business optimisation and automation like autonomous cars have benefited from these new Artificial Intelligence (AI) methods. Some of these ML models are used in ways that their predictions can affect people's safety or commercial success. These models must be considered trustworthy with errors detected {\color{black}and dealt with before they can affect the success or safety of the process being controlled.}

Neural Networks (NNs), and in particular Deep Neural Networks (DNNs), represent one such class of ML algorithm. Due to the nature of DNNs, the decisions they produce can seem arbitrary. {\color{black}These DNNs are comprised of thousands of nodes that perform mathematical operations, creating a "black-box like" system, in which one is unable to judge the decisions being made by simple inspection. This is especially true for the operators of these networks who may not have the technical knowledge of how such models work. The researchers and engineers who build such models} understand the underlying mathematics, but that does not give enough insight into how the model will produce a decision without further analysis. The effort to explain the decisions of DNN has become known as Explainable Artificial Intelligence (XAI). XAI techniques have varied in methods and applications {\color{black}as the type of explanation needed is dependent on the system and model used.} This paper will review how the different methods are applied across industries to try and gain insight into the path that XAI is going to take.

\begin{table}[b]
    \centering
    \rowcolors{2}{gray!40}{gray!20}
    {\color{black}\begin{tabular}{p{1.3cm}p{5.6cm}p{1.3cm}p{5.2cm}}
        \hline
        Acronym & Title & Acronym & Title \\
        \hline
        A3C	&	Asynchronous Advantage Actor Critic	&	LMT	&	Linear Model Trees	\\
        AI	&	Artificial Intelligence	&	LMUT	&	Linear Model U-Trees	\\
        ARS	&	Adaptive Region Scoring 	&	LRP	&	Layer-wise Relevance Propagation	\\
        CEP	&	Complex Event Processing	&	MAS	&	Main Aggregator Server	\\
        CNN 	&	Convolutional Neural Networks	&	MCRTS	&	Monte Carlo Regression Tree Search	\\
        COGLE	&	Common Ground Learning and Explanation	&	MDL	&	Minimum Description Length	\\
        CSG	&	Curios Sub-Goal	&	MDP	&	Markov Decision Process	\\
        DDDQN	&	Deuling Double Deep Q-learning Network	&	ML	&	Machine Learning	\\
        d-DNNF	&	deterministic Decomposable Negation Normal Form	&	Mo{\"E}T	&	Mixture of Expert Trees	\\
        DDPG	&	Deep Deterministic Policy Gradient	&	MXRL	&	Memory-based Reinforcement Learning	\\
        DDQN	&	Double Deep Q-learning Network	&	NLDT	&	Non-Linear Decision Trees	\\
        Deconvnet	&	Deconvolutional network	&	NN	&	Neural Networks	\\
        DNN 	&	Deep Neural Networks	&	PER	&	Prioritised Experience Relay	\\
        DNTS	&	Distance to Nearest Training Sample	&	PG 	&	Policy Graphs	\\
        DQN	&	Deep Q-learning Network	&	PPO	&	Proximal Policy Optimisation	\\
        DRL	&	Deep Reinforcement Learning	&	RAMi	&	Represent And Mimic framework	\\
        DT	&	Decision Trees	&	RNN	&	Recurrent Neural Network	\\
        ETeMoX	&	Event-driven TEmporal Models for eXplanation	&	SARSA	&	 State-Action-Reward-State-Action	\\
        GAN	&	Generative Adversarial Agent	&	SDRL	&	Symbolic Deep Reinforcement Learning	\\
        GNC	&	Guidance, Navigation and Control	&	SHAP	&	Shapley Additive exPlanations	\\
        GRAD-CAM	&	Gradient-weighted Class Activation Mapping	&	TM	&	Temporal Model	\\
        GTr	&	Gated Transformer	&	UAV	&	Uncrewed Arial Vehicle	\\
        HVAC	&	Heating, Ventilation and Air Conditioning	&	VEE	&	Value Estimated Error	\\
        IB	&	Information Bottleneck	&	VR	&	Virtual Reality	\\
        LIME	&	Linear Interpretable Model-agnostic Explanations	&	XAI	&	eXplainable Artificial Intelligence	\\

        \hline
    \end{tabular}
    \caption{The acronyms used throughout this review.}}
    \label{tab:my_label}
\end{table}

Deep Reinforcement Learning (DRL) is a type of ML that, instead of using a data set to train the model, it employs a Markov Decision Process (MDP) that learns through trial and error. The model will choose decisions by trying to maximise a reward function. Over many decisions, the algorithm will learn which actions produce the highest rewards and this will eventually lead to an optimal solution. {\color{black}Such solutions found by DRL models may coincide with solutions or strategies implemented by a human operator. This is seen when a DRL agent is used to play the Atari game Space Invaders where the strategy for both humans and DRL is similar\cite{stamper2019exploring}.} However, this method has the benefit of being able to find innovative solutions, leading to strategies that will differ from human operators. {\color{black}These novel strategies can come in many forms such as by exploiting the game mechanics or taking risks that human players would not take.} As there is no guidance apart from a simple reward function, the model is able to find the most efficient solution.

As the DRL method for creating models can lead to more innovative solutions that are not obvious to a human observer, it is important that XAI can be used to check the solutions are correct. These explanations can also be used to examine any novel strategies or solutions that these DRL models formulate, such as learning why a strategy in a video game has been chosen. As DRL models are used in time sensitive applications where a decision is made at every time step, explanations must be delivered promptly to the user. Any safety critical operation of a model will require the trust of an operator with no knowledge of the model being used, therefore the explanations have to be understandable to be able to build that trust. {\color{black}This level of explanation can change with the system being operated, as an operator observing the AI control of a production line will be more knowledgeable about their system than a member of the public in an autonomous car.} With these applications that are safety dependent, any accidents that take place under the control of a NN, XAI can be used to analyse the post-accident data to determine the fault. Being able to access the fault that has caused the accident could help with preventing future accidents or help with any insurance claims.

This review will first give an overview of the two subjects, DRL and XAI. After this, a discussion of the state of the field as it stands at the moment will be done with a look at the work of two previous reviews. Following this will be the selection criteria that is used to choose the papers to be reviewed. After this, there will be a summary of each paper before some concluding remarks on the limitations and future areas to study. A list of acronyms and initialisms can be found in Table 1.

\section{Deep Reinforcement Learning}
The original method for applying deep learning methods to reinforcement learning methods was the application of Q-learning formulated by Mnih et al.(2015)\cite{mnih2015human}. The Q-learning process begins with a MDP that takes the variables \\($S, A, R, P, \gamma$) to interact with the environment, where we have: 
\begin{itemize}
    \item S: Set of the States.
    \item A: Set of the Actions.
    \item R: Set of all the possible Rewards from the actions taken. The rewards are denoted as being $R_{t+1}$ after the action $a_t$ and state $s_t$. The reward can also be described as $R(s)$ where this is the reward for reaching state $s$.
    \item P: Transition dynamics where $P(s'|s,a)$ is defined as the distribution of the next state $s'$ where an action $a$ has been taken from state $s$, where $s,s'\in S, a\in A$. The transition dynamics at the beginning are denoted as $P(s_0)$ or $\rho_0$.
    \item $\gamma$: The discount factor with range $[0,1]$.
\end{itemize}

To solve the MDP all these values need to be known, but in many cases, the reward function and the transition dynamics are not known. To solve this problem, trial and error is used to explore the environment to build an understanding of the inner information. This process is known as reinforcement learning. The aim of reinforcement learning is to map the policy $\pi$ between the states and actions that maximise the expected discounted total reward over the agent's lifetime. The action value function is also known as the Q function which describes this relationship as: 
\begin{equation} \label{eu_eqn}
  Q^{\pi}(s,a)=E^{\pi}[\sum^T_{t=0}\gamma^tR(s_t,a_t)]  
\end{equation}
In this equation, $E^\pi$ is the expected value by following policy $\pi$. This Q function can be solved using the Bellman equation:
\begin{equation} 
    Q^\pi(s,a)=R(s_t)+\gamma E[Q^\pi(s_{t+1},a_{t+1})]
\end{equation}
This equation allows for the Q-learning and State-Action-Reward-State-Action (SARSA) recursive estimation procedures. It is possible to then improve the policy from $\pi$ to $\hat{\pi}$ by greedily choosing $a_t$ in each state:
\begin{equation} 
    \hat{\pi}(s_t)=argmax_{a_{t}}Q^\pi(s_t,a_t)
\end{equation}
When paired with the use of DNN that can perform as a function approximator, this technique has led to the increase in the use of DRL for complex tasks. These DNNs are also the reason the explainability problem that needs to be solved using XAI.

As DRL has matured there have been several innovations in the field that have created more efficient methods for training and better NN architectures that can be used in a wider variety of situations. The field is split into model-based DRL and model-free DRL. Model-based systems use a defined model that can be queried to provide future rewards and states allowing for the DRL agent to plan ahead to maximise the reward in the future. This was used by Google's AlphaZero model (Silver et al.(2018)\cite{silver2018general}) to be able to strategise many moves in advance. In these model-based DRL modes, the model can either be learned or given to the agent.

In model-free agents, the process skips learning the model and just learns the policy and makes decisions based on previous actions rather than predicted rewards. These types of models have become more prevalent as they are easier to set up and can be used in a wider variety of areas. These agents come in three types: those that utilise Q-learning such as Double Deep Q-learning Network (DDQN), those that use policy optimisation such as Proximal Policy Optimisation (PPO) or Policy Gradient, or those that use a mixture such as Deep Deterministic Policy Gradient (DDPG) and its derivatives. Most papers in this review will be using one of these types of DRL as they are more widely researched.

\begin{figure}
    \centering
    \includegraphics[width=10cm]{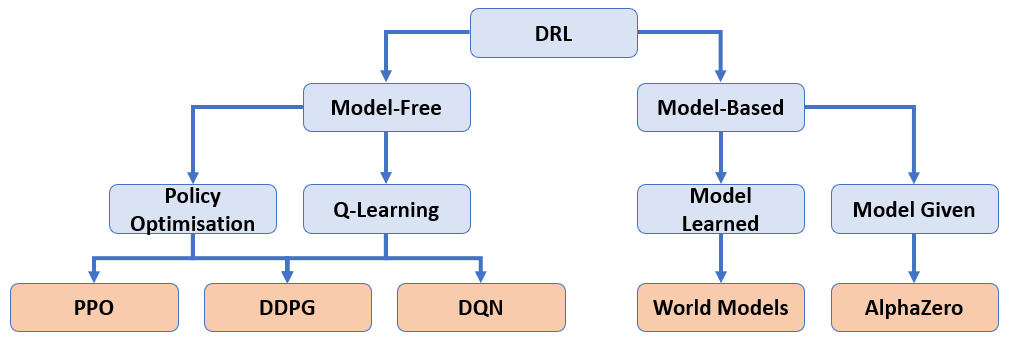}
    \caption{The types of DRL schemes with some examples.}
    \label{fig:my_label}
\end{figure}

{\color{black}There are wide ranging applications for DRL agents and many of them are featured in this review as there have been attempts to explain their decisions using XAI. These are areas such as video games, vehicle guidance and navigation, systems control, manipulator control, network monitoring, and medical diagnosis. There are however some applications of DRL that have not exploited XAI methods to enable greater interpretability; these are areas such as audio-based reinforcement learning, emotion recognition using DRL, and phonetic text enrichment using DRL. These research fields will not be featured in this review as no papers meeting the parameters set out in the selection criteria exist.}

\section{Explainable Artificial Intelligence}
{\color{black}The problem of not understanding the ML models being produced has plagued researchers since their inception.} As such there has been a race to develop tools that can explain model behaviour. Explainability has been a part of AI research since the 1970s with approaches such as decision trees (DT), though it has risen to importance in the last decade as ML has become {\color{black}more prominent and ever more opaque. This rise in opaqueness is due to the AIs become more capable the NN also become more complicated and difficult to decipher.} Therefore, the methods used to explain AI models have had to evolve.

 {\color{black}Explainability is using processes to help make an AI model more transparent to the user by producing explanations. These explanations describe the actions or decisions that an AI may have taken. Explainability may also be an AI system that can be questioned or visualised, though this review will focus on the former definition.}

XAI methods can be categorised depending on whether they describe part of the model or the whole model and when they provide an explanation. Explanations can be intrinsic or post-hoc. Post-hoc is where the explanations are derived from an already trained model and can be model-agnostic or model-specific. Intrinsic explanations are when the model is already explainable such as a decision tree. These types of explanations are always specific to that model. The scope can either be local or global. The global explanations will explain all the model's actions, while a local explanation will only explain one action.  

\begin{figure}[H]
    \centering
    \includegraphics[width=8cm]{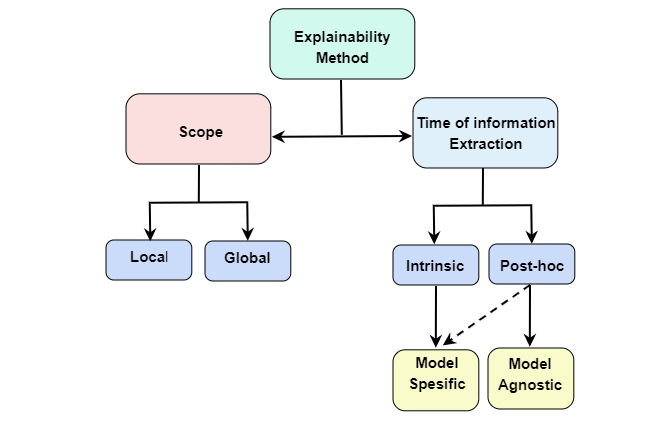}
    \caption{The two branches that define the type of explainability method.}
    \label{fig:explainability tree}
\end{figure}

The methods of XAI fall into a few different categories. The first is to create symbolic DTs that are intrinsically explainable, which learn to mimic the actions of a more complex network such as a DRL model. The second type is for models that have a visual input. Saliency maps show areas of interest to the model and thus affect the decisions the model takes. The third is to assign values to the model inputs to find how important it is to the output using mathematical methods. The two most frequently used are Linear Interpretable Model-agnostic Explanations (LIME) and SHapley Additive exPlanations (SHAP). {\color{black}There are further methods but these are three common types. The review will look at examples of these three methods and the less common methods of XAI and their applications throughout the review.}

\section{State of the Field}
Recently the publication rate of XAI papers has increased exponentially. The methods and areas that XAI is applied to are in constant motion. It is paramount for research into XAI to apply the correct DRL application. It is vital that XAI methods are understood and that the implementation by research teams is appropriate. It is imperative to keep an overview of a research field undergoing a rapid rate of advancement.

\subsection{Previous Reviews on the use of XAI in DRL}
There have been reviews done previously that cover this subject. The first of two to be looked at is by Wells and Bednarz (2021)\cite{wells2021explainable}. They reviewed 25 papers published between 2017 and 2019. Most of these papers were from 2019 as the rate of studies covering both XAI and DRL rose quickly over the selected years. The authors looked at five different areas of XAI: human collaboration, visualisation, policy summarisation, query-based explanations, and verification. Rather than focusing on the methods used to generate the explanations, the authors looked at the types of explanations that were given. Due to the papers being selected from an earlier period they do miss out on some new methods that have become important in the XAI field. These include LIME and SHAP, which have become heavily featured in the papers covered in this review. This highlights the fast pace at which this field of study is moving. The authors did not look at the fields in which these XAI methods have been applied. {\color{black}This is an area that our review looks to explore.} Their review concluded that there was still a way to go to make DRL agents interpretable. They found that an issue with XAI solutions shown in the papers they reviewed is the demonstrations were either on simplistic models or lacked scalability. Having explanations for models that have no bearing on real world problems are not as useful. On the other hand, they did point out that research in this area is in its infancy so simple models are to be expected. The authors also were disappointed with the lack of testing. When there was testing it was of limited scope. Their final major conclusion was that the explanations ended up giving too much information. This over saturation of information can lead to confusion about what the explanations are suggesting.

The second review is by Vouros (2022)\cite{vouros2022explainable}, who looked at four problems affecting interpretability in DRL and then used papers to propose solutions to those problems. The problems they suggested were model inspection, policy explanation, objectives explanation, and outcome explanation. Their review covered 31 studies in depth with most papers coming from 2019 or 2020. This review also did not focus on the applications for explainability in DRL. This review found that it is difficult to quantify what constitutes a good explanation as there are no studies looking at this in relation to AI. They have a list of suggestions that should be implemented to increase the level of interpretability. They include building an XDRL toolbox to give explanations that are comprehensive as needed as well as transparent. This needs to be done in conjunction with defining interpretability, explainability and transparency.

\section{Problem Statement}
In light of the previous explanations of XAI and DRL in the preceding sections, the case has been made for maintaining a proper understanding of how XAI is being applied to DRL methods. This review will look at the current fields in which researchers apply XAI to DRL models, what implementations of XAI researchers are using, how practical are these implementations of XAI, and what are the shortcomings of these studies. 

\subsection{Selection criteria for review papers}
To select the papers to review, Google Scholar is used to search for the most recent academic papers that included Explainability and DRL as the main focus of their studies. The search parameters used the terms "Explainability", "Interpretability", "XAI", "DRL", "Deep-Q", and "Deep Reinforcement Learning". Papers from the previous three years were preferred to limit overlap with other reviews. The selection criteria were as follows: the NN should include an XAI method to explain parts or the whole of a decision made by non-intrinsic AI networks, and the NN must be a DRL network. This is a necessary distinction since some studies use a DRL network to generate their explanations. These XAI methods are not in this review's purview.

{\color{black}Unfortunately not all fields of research DRL models are used, such as audio processing, can be included in this review as they did not contain the XAI work that was needed to meet the criteria. The same can be said for XAI methods that might not be included because no papers applied them to a DRL model and therefore did not meet the strict limits of this review.}

A total of sixty-two papers were initially selected for review. These papers were classified by the field in which they apply. Three research papers were discarded for either being, withdrawn by the authors or found to provide little contribution to the topics needed.

The reviews done before into XAI and DRL have usually split the papers into categories based on the type of XAI used. This review instead looks to categorise based on the application the study is using the DRL in. Categorising in this manner will allow for an analysis in which areas certain types of XAI methods excel. It will also highlight where there needs to be more research done in applying XAI to DRL methods or implementing a particular XAI technique in a research field.

\section{Research Fields with XAI applications}
The applications of XAI in DRL found when gathering the relevant papers are as follows; playing video games, vehicle guidance, systems control, robotic manipulators, network solutions, and medical applications. The most prevalent use of XAI with DRL is that which uses a video game to test the use of explanations. {\color{black}These video games provide a non-compute intensive method to train and test a DRL model quickly as the simulation speed can be increased. The games are also simple to adapt to a DRL type NN as the reward function can use the in-game score to judge success. There are also defined rules that applied to the player in such games which allows for set boundaries for DRL agents and easier extraction of explanations. These video games also allow for the altering of the difficulty which allows analysis of how DRL agents will cope with a changing challenge. With some video games being more complex, multiple strategies can be tried to maximise the in-game score. The DRL agents can even be used to find novel strategies that out-perform human strategies.}

The second field that has seen the adoption of the DRL scheme and the use of XAI is the use for guidance, navigation and control (GNC) tasks. These GNC tasks cover multiple vehicle types, but as they share commonalities they have been combined into one subject. The types of vehicles seen are Uncrewed Aerial Vehicles (UAVs), automobiles and ships. UAVs were the most common vehicle type in the papers reviewed. With affordable off-the-shelf models of UAVs available, real-life testing is possible for most research teams. There are also good software packages that allow research teams to simulate UAVs, which is most useful when training NN. Autonomous ground vehicles were the second most common application in this field, varying from small battery-powered wheeled robots to full-sized cars. This variation allows a low barrier of entry on the smaller end. The larger vehicles are most representative of a high growth market. With this application, there are many datasets being created to train DRL models. Simulations also make these vehicles highly accessible. The final modality is that of autonomous ship control. However, there is only one paper that deals with nautical operations but this could be an area of interest in the future. The need for explanation is vital in this field as these applications, especially autonomous cars, are the most likely to interact with the general public and thus need to garner trust to be accepted for use.

The third field is that of system control. System control covers topics such as power system management and traffic light control. These are critical systems in which knowing the reasons behind a decision taken by a DRL model that is running them is vital to maintaining safety and increasing efficiency. {\color{black}The explanations for this field will be more technical than those of a public facing solution as the operators are more likely to be trained in the use of these models.} The explanations may require to be generated in real-time. This means computing time may become an important factor.

The fourth field is for Robotic Manipulators. Studies in this field look at how XAI can increase the understanding and performance of DRL controlled manipulators. Robotic manipulators have seen usage in manufacturing since the 1960s. These robots, however, had a set action from which they could not deviate. The new DRL controlled manipulators, on the other hand, can adapt to a variety of tasks or non-perfectly aligned tasks. As the technology grows into more assembly lines the agent's behaviour must be known for safe operation.

The fifth field is using DRL networks to optimise mobile network solutions. Only a couple of papers covered this topic in the period for review, which suggests this is a new field to exploit. It is imperative to build understanding as these networks are so vital to how modern life runs. Knowledge of bandwidth usage in certain situations can allow engineers to construct redundancies into the correct parts of the networks.

{\color{black}The final section covers the single paper that didn't fit into the other sections. The paper applies their XAI method to a few different applications including the only medical application seen in any of the reviewed papers.}

\begin{figure}[h!]
\centering
\includegraphics[width=\textwidth]{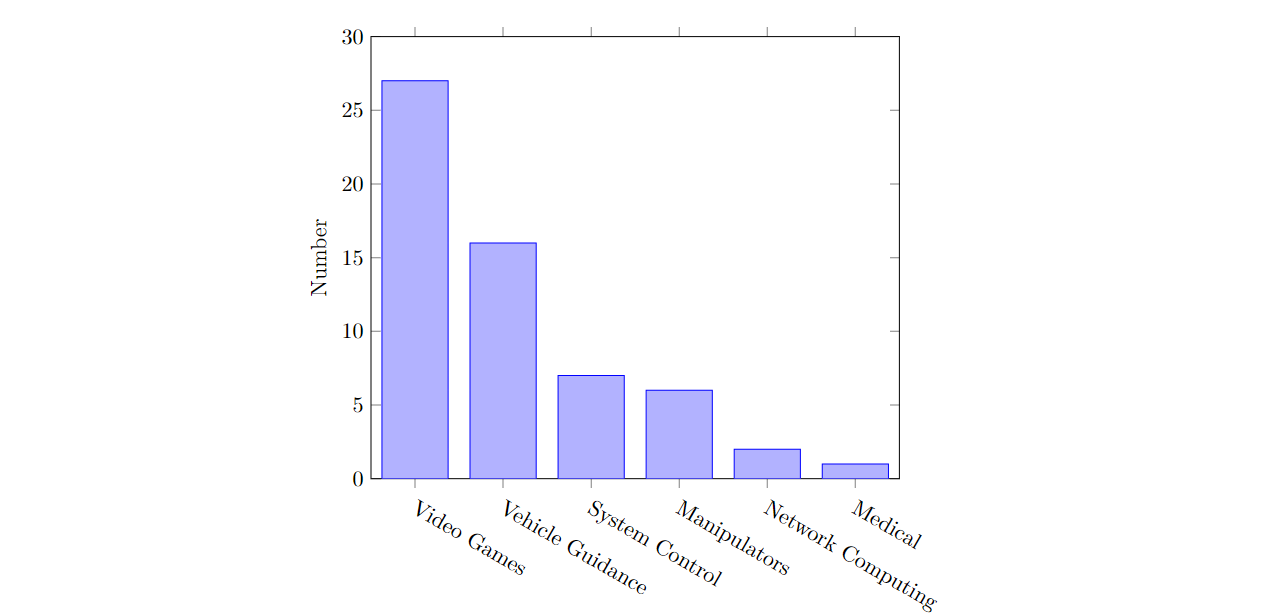}
\caption{Areas of Studies for the Reviewed Papers}
\end{figure}

Of the 59 studies chosen for this review, 27 of them were studies that used a video game to train their networks and run their tests on explainability. The second most common was the study of XAI in Vehicle Guidance with 16 papers reviewed. The third most common with seven research papers were studies using XAI in control systems such as traffic light controllers. Next, there were six papers written on the use of Robotic Manipulators. Two papers were reviewed on the application in the field of Networks. The final paper covers the medical applications of XAI in DRL. 

{\color{black}Just under half of the papers were published in 2021 with 26 papers. 10 papers were from 2019, 11 from 2020, and 12 were from 2022. The large jump from 2020 to 2021 could be due to a slump in publishing rates from the effects of the Covid pandemic. The relatively small amount of papers from 2022 is because the year wasn't fully complete when the papers were collated.}

\subsection{Video Game Simulations}
Video games are a good testing ground for XAI techniques, so there is no surprise that there are many methods to be reviewed. These include mathematical methods where the model inputs are varied to find the effect on the output like Shapley values; the use of symbolic networks to emulate a trained DRL network where the DT itself is explainable; the use of visualisation techniques such as saliency mapping to see where the agent is focusing its attention; finally, methods that use separate NN or added layers to the agent network to generate explanations. This section will go through the different XAI techniques proposed by these papers to show their implementation.

The six papers that used visual explanation to describe the actions of the DRL network are first covered. Beginning with the work of Joo et al.(2019)\cite{joo2019visualization}, the authors described the Gradient-weighted Class Activation Mapping (Grad-CAM) system for classification tasks that use Convolutional Neural Networks (CNN). They apply this Grad-CAM to the Asynchronous Advantage Actor Critic (A3C) DRL developed by DeepMind and run through a series of Atari video games.

\begin{figure}[H]
    \centering
    \includegraphics[width=6cm]{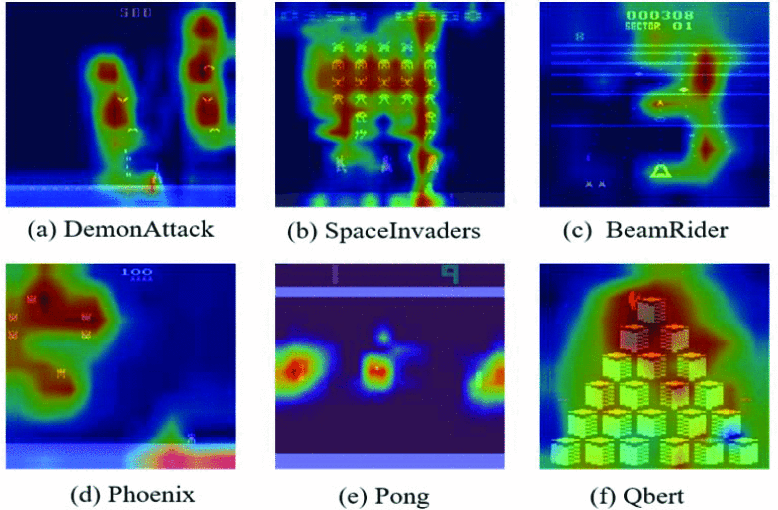}
    \caption{Grad-CAM images generated}
    \label{fig:Grad_CAM}
\end{figure}

The team managed to create Grad-CAM results for various Atari games shown in Fig. 4. The results generally show that the focus of the DRL agent is on the more crucial parts of the screen. In the paper, they conclude that the use of the Grad-CAM could lead to insights for expert users rather than the general public due to the knowledge needed to understand the explanations. They suggest these explanations will allow a deeper understanding of how DRL agents learn and operate.

Next, the work by Douglas et al.(2019)\cite{douglas2019towers} is reviewed. Their paper uses a saliency map generation technique and then uses that to create understandable visualisations for the end-users. The video game they used to train their agent was the Pommerman benchmark domain. They had to adjust the usual saliency algorithms by moving from a pixel by pixel method to focusing on each game square instead. The second modification was to record the change in the saliency values produced by each action, normalising these changes to between zero and one. The result was a compromise between emphasising changes in magnitude and changes in positive and negative values.

\begin{figure}[H]
    \centering
    \includegraphics[scale = 0.5]{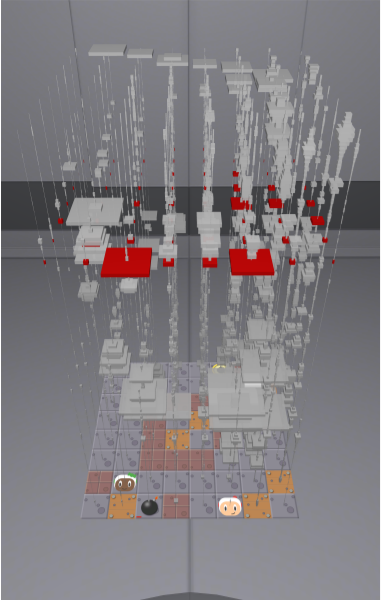}
    \caption{One of the generated "Towers of Saliency", the game board shows the current positions and the highlighted red areas are the current Saliency map.}
    \label{fig:Tower of Saliency}
\end{figure}

These saliency maps were then taken for a whole game and arranged into a 3D saliency map that the paper has taken to call "Towers of Saliency". With these towers, the visualisation of the progress of the importance of the game squares is achieved, these visualisations can be seen in Fig. 5. A Virtual Reality (VR) headset can be used to examine the results, which allows for an innovative view that can help with understanding. Overall this paper generated a visualisation method for use in video games where each game is over a short time frame. The visualisation is then able to show the entire progress of the game. The authors believe that the system can drive a greater understanding of the field.

Next, the paper by Guo et al.(2021)\cite{guo2021machine} looks into the differences in attention between human players and DRL agents. They do this by producing saliency maps for the AI players while tracking the eyes of the human players. They found that the attention maps became more similar as the agents trained. The similarity of the attention maps was useful for predicting the model's success. Their work also looked at discount factors and found longer-term outlooks that are too long can lead to the model becoming too distracted by long term goals. They also found that these maps could be used to detect the reason for a failure state, either by the model having its attention on the wrong object or making the incorrect decision. Finally, they found that this could extend to many different types of DRL agents. 


In the next paper, Anderson et al.(2019)\cite{anderson2019explaining}, looked at explaining the DRL with the use of Saliency maps and reward-decomposition bars. They tested these explanations with 124 naive human participants and found that they needed both the saliency maps and reward-decomposition bars to gain an understanding of how the agent was working. They also found that different situations and different people required different explanations to understand the models. Therefore they found it is best to give multiple explanations to cover more bases.

In Dao et al.(2021)\cite{dao2021learning}, the authors used Grad-CAM to visualise snapshots in Atari video games. They then analysed how many snapshots are required to understand the DRL agent. They did this by utilising state-value approximations to construct sparse-bias space. They reduced the number needed for analysis by grouping states holding similar attention in the image. They found that these reduced the number of snapshots required for interpretation. The selection of snapshots chosen to be interpreted or discarded gave insight into how the model behaved.

The following paper for review is by Huber et al.(2019)\cite{huber2019enhancing}. The authors used a variation on the Layer-wise Relevance Propagation (LRP) scheme. This scheme uses only the most relevant neurons in the CNN to generate saliency maps that highlight the paramount areas for the agent's decision making. Using a Dueling Double Deep Q-learning Network (DDDQN) they tested the scheme on three Atari games and found that the new scheme managed to highlight more pertinent information on the screen. The authors were able to show that this scheme worked on the most current versions of the Deep Q-learning Network (DQN) algorithm.

{\color{black}Liu et al.(2022)\cite{liu2022towards} look to improve the training efficiency and performance of DRL methods as well as providing explanations with their Adaptive Region Scoring (ARS) mechanism. The mechanism produces scoring maps that show the important areas of the screen like a saliency map. This map is then used as an explanation for the user as well as guiding the DRL agent to the areas that it needs to focus upon. The authors found that the addition of this module managed to improve the quality of policy of learning across a range of modern DRL methods. This paper is a good example of using XAI methods to improve the performance of DRL models.}

Next, the paper by Druce et al. (2021)\cite{druce2021explainable} produces three explanations for the actions taken by the agent. The three explanations are through a graphical depiction of the performance in the game state, a measure of how the agent would perform in similar environments, and a text-based explanation of what these two explanations imply. The Value Estimated Error (VEE) measures how well the agent estimates its state. A lower value means that the agent has seen a similar state in training. The second value is the Distance to Nearest Training Sample (DNTS). They created a user interface (UI) for this that shows the various explanations, shown in Fig. 6. 

\begin{figure}
    \centering
    \includegraphics[scale = 0.55]{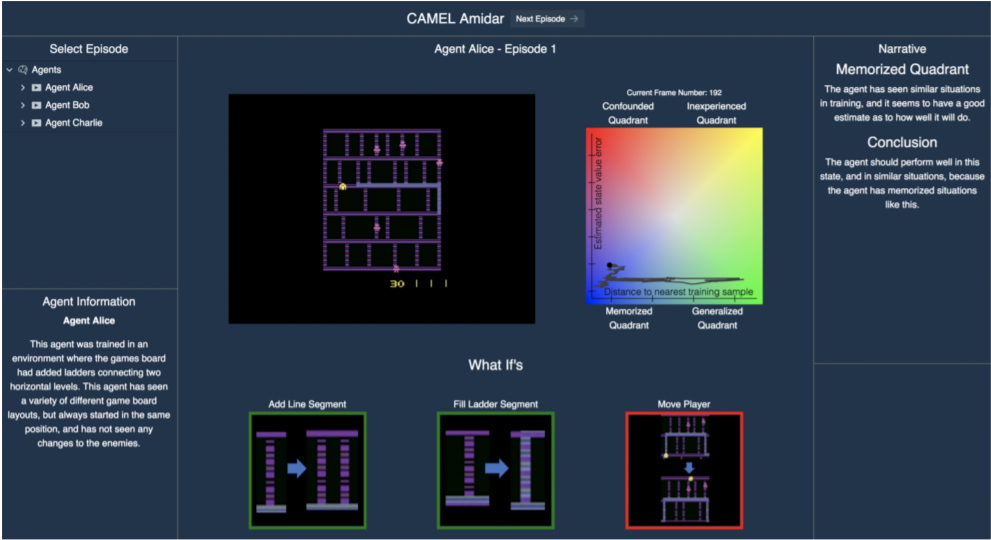}
    \caption{The UI used to show the different types of explanation.}
    \label{fig: The UI}
\end{figure}

The visualisation was tested on the Atari video game Amidar, with participants in the experiments shown several scenarios and then asked about how they trusted the agent. The study found that their visualisation software increased trust in the agent. The authors were disappointed that the trust only rose by one standard deviation. 

The following study by Druce et al.(2021)\cite{druce2021brittle} looked at using human-machine teaming for the AI to work in conjunction with a human teammate and then use explanations to justify the AI's actions. The authors used the StarCraft 2 video game to run their DRL agent. They used CAMEL to generate the explanations and then show them to the participants. They found that the participants gave flawed explanations for the AI behaviour by adding human-like motivations, such as the AI getting scared of certain enemies. They also found that when a reasonable explanation was not producible by CAMEL it was because the agent was acting unreasonably.

The next few papers move away from the use of visualisation to explain the agents and instead use modified NN architectures to increase the explainability of the networks. The first of these is by Lyu et al.(2019)\cite{lyu2019sdrl}. They leveraged symbolic planning to enable the DRL model to maintain its high-dimensional sensory inputs while enabling task-level interpretability. It does this by linking the symbolic actions to options and using the architecture to create subtasks that the DRL learns and a meta-controller that suggests new rewards. They found the Symbolic Deep Reinforcement Learning (SDRL) framework successfully utilised symbolic planning to improve task-level interpretability.

In Guo et al. (2021)\cite{guo2021edge}, the authors proposed a novel self-explainable model EDGE that modifies a Gaussian process with a customised kernel process and an interpretable predictor. They also developed a method to improve learning efficiency using a parameter learning procedure that leverages inducing points and variational inference. Testing these solutions on Atari and MuJoCo games, they found that they could form strategy level explanations by predicting final rewards generated by the agent and extracting time step importance in the levels. They found that using the explanations generated for several tasks like discovering vulnerabilities or errors in the policy was possible. They also found that this scheme could defend against adversarial attacks.

Landajuela et al.(2021)\cite{landajuela2021discovering}, came up with a deep symbolic policy to help with the interpretability of DRL agents. This approach directly searches the space for symbolic policies. Using an autoregressive Recurrent Neural Network (RNN) to generate the control policies that can be represented by tractable mathematical expressions. To maximise the performance of the generated expressions they used a risk-seeking policy gradient. Across eight video game tasks, the symbolic policies outperformed the state of the art DRL schemes they tested against while being readily interpretable. The authors felt that this is a viable alternative to NN when working with systems that need to be interpretable but have deployment constraints such as memory or latency limits.

The following paper is by Sieusahai et al.(2021)\cite{sieusahai2021explaining}. They used an interpretable surrogate model that predicts how the primary model acts. This surrogate model works by transforming the pixels that are input into the DRL model into an interpretable, percept-like input representation. They found that their trained model could accurately predict the actions of the primary model on a wide range of simple 2D games at about 90\%. The authors decided to show both models' adversarial examples to see if the surrogate model would predict the right action. This experiment showed that the surrogate model changed the same way as the primary model showing that they used the same features to find the following actions.

In the following paper, Barros et al.(2020)\cite{barros2020moody} used the Moody framework to explain the behaviour of a DRL agent in a competitive card game. The Moody framework tries to create a representation of the agents by using the Pleasure/Arousal model. The Moody framework works in competitive environments by measuring the confidence of the opposing player after the agent's actions. These confidence measures transform into a pleasure/arousal scale representing two scales pleasing to unpleasing and excited to calm. The authors found that this framework gave the model an enriching explanation of its actions. They also managed to assess the accuracy of understanding the confidence of the opposing player.

In Climent et al.(2021)\cite{climent2021applying}, the authors used Policy Graphs (PG) to create a framework for explanations. These PGs represent the agent's behaviour as this team tries to measure the similarity between the explanations and the actions. The policy graph is a generated list of statements that describe the system's state. In this case, its usage is to balance a virtual cart pole. They found that controlling the actions using the policy graph they created almost reproduced the success of the DRL agent at controlling the cart pole. The success rate at holding the cart pole vertically did drop by a small amount but can still be considered a success. More research is required as this study was only limited to one example.

In the following paper by van Rossum et al.(2021)\cite{van2021novel}, the authors used a Curios Sub-Goal focused agent (CSG) to attempt to explain a DRL agent. The agent is split into three parts. First, they used a Generative Adversarial Network (GAN) to generate the next state. This GAN can build an understanding of the object interaction and actions the agent takes within the environment. The second part was the traditional DRL agent that dealt with player navigation. For the final section, the team broke down the task into easily understood sub-goals that allowed for easier understanding than trying to understand the whole strategy. The authors found that this method successfully broke down the tasks into understandable sub-goals and allowed for a greater understanding of how the agent behaved. They did note that this was a simple task limited to discrete observation spaces with basic dynamics. They did speculate that a more advanced DRL could be integrated and suggested this for future work.

These papers take the DRL agents and try to use DT to emulate them and turn the black-box agent into an interpretable model. In Ding et al.(2020)\cite{ding2020cdt}, the authors used cascading DT to accomplish a DRL agent in an explainable way. The use of DT is to learn the actions of a DRL agent and mimic them. DT are inherently interpretable as the rules are easy to follow along the nodes. This paper suggests using Cascading Decision Trees (CDT), which cascades a feature learning DT and a decision making DT into one model. The CDT has a better function approximation in both training and operation. The reduced number of parameters while keeping the tree prediction accuracy was also achieved. The authors also found that while the CDTs had good model approximation in the imitation learning setting, they did struggle with instability in different imitators in the tree structures that affected interpretability.

The authors of the following paper Vasic et al.(2022)\cite{vasic2022moet}, used a Mixture of Expert Trees (Mo{\"E}T) to produce verifiable reinforcement learning. Mo{\"E}T is a model that consists of decision tree experts and a generalised linear model gating function. The new scheme allows decision boundaries with hyperplanes of arbitrary orientation instead of the usual axis-perpendicular hyperplanes seen with other schemes. The authors also designed a scheme that they called Mo{\"E}T\textsubscript{h} that bases decisions on a single expert chosen by the gating function, this change allows for easy decomposition into a set of logical rules to generate explanations. The authors found that this scheme outperformed other decision tree techniques in model reproduction and still produced the means of model verification.

Liu et al.(2021)\cite{liu2021learning}, suggest creating a Represent And Mimic framework (RAMi). This model can capture the independent factors of variation for the objects using identifiable latent representation, using a mimic tree that measures the impact on the action values by the latent features. A novel Minimum Description Length (MDL) objective based on the Information Bottleneck (IB) principle was used to optimise the tree's fidelity and maintain its simplicity. They used a Monte Carlo Regression Tree Search (MCRTS) algorithm to find an optimal decision tree. These optimised trees showed strong approximations with fewer nodes than the baseline models. The interpretability was tested by showing latent traversals, decision rules, causal impacts and human evaluation results. In the human evaluation, they found that 83\% of respondents preferred the mimic tree solution for interpretability compared to saliency and superpixels.

In the following paper, Dhebar et al.(2022)\cite{dhebar2022toward} use Non-Linear Decision Trees (NLDT) to provide interpretability for DRL agents. They build on previous work using NLDT by introducing an evolutionary computation function to optimise the decision tree. Data from a pre-trained DRL model for the moon lander game is used to train an open-loop version of the NLDT. Training is done using a recursive bilevel evolutionary algorithm. The top part of this open-loop NLDT is then trimmed off and then trained using an evolutionary optimisation in a closed loop. This process gives the final NLDT to be used. This method created an interpretable model that operated on discrete action problems. This study did not go into the interpretability of the model and just relied on the inherent interpretability of DT.

{\color{black}Luan et al.(2022)\cite{luan2022lrp} developed a method for condensing a DRL model to save memory and computation resources for deployment. They used the XAI method LRP to trim down their model to maintain its performance while improving its efficiency. This was accomplished by using the relevancy scores provided by the LRP. The CNN layers of the model were ranked and pruned using the relevancy scores along with fine tuning accomplished with policy distillation. The new efficient model was then tested in a variety of Atari games to validate its performance. Their pruned model was able to outperform other methods for pruning models across all three Atari titles tested but particularly the BankHeist and RoadRunner games. This paper stands out as the XAI method is being utilised to provide more value than just explaining model decisions. }

Liu et al.(2019)\cite{liu2019toward} proposed using Linear Model U-Trees (LMUTs) to approximate the DRL predictions. The team put forward a novel training method for the decision tree where the mimic learner observes the ongoing interactions between the environment and the DRL agent using an online algorithm. U-Trees are a natural choice for modelling DRL agents as they are essentially regression trees for value functions. These are slow in training leading the team to use an LMUT that allows the nodes to be linear models. This linearity allows for a better approximation and a smaller tree that is better for interpretation. They tested on three simple games and found that their LMUT could approximate the Q value function learned by the DRL, meaning that the black box DRL became more interpretable as a result.

Finkelstein et al.(2021)\cite{finkelstein2021deep} use a taxi navigation game to look into the use of DT to help explain the actions of a DRL agent. They propose to use MDP transforms to explain the behaviours. These transforms to the MDP are searched for examples where the actor's behaviour is as expected. If the change from one agent to the new agent is meaningful, the resulting change is identified as an explanation. The team found that these model transforms produced interpretable explanations. 

In Jaunet et al.(2020)\cite{jaunet2020drlviz} the authors looked at the memory that the DRL agent was accessing to determine the decisions that the agent was taking; they named this system DRLViz. The memory stores a vector of a timestep that can be viewed using their developed software, which can be compared to other situations and similar vectors can be seen for similar situations. These explanations are limited to expert users as a naive user would gain no insight from the memory data. To validate, they consulted three experts from the field to give their thoughts on the DRLViz package, who stated that it provided good explanations; in particular, two out of the three thought the software was easy to navigate while the third took some time to become comfortable with it. Overall, the authors developed a tool to look at explanations for decisions by DRL agents in a novel way. 

The next paper by Karalus et al.(2021)\cite{karalus2021accelerating} uses counterfactual explanations along with human-in-loop training to try to expedite the training of the agent. In their research, they use TAMER, which replaces the automatically generated reward in the training of the DRL with a human-generated reward. Specifically, the authors employ the DeepTAMER architecture that applies a DNN to approximate the H function that is used in turn to approximate the human-based reward. The counterfactuals are generated by suggesting an action to be taken by the agent if some other input is changed in some way, these counterfactuals were only shown during negative rewards to teach the agent how to move to positive reward scenarios. In their experiments they found using this scheme did increase the training rate seen especially during the early stages of the training when there were large differences between the actions and the counterfactuals.

In Cruz et al.(2019)\cite{cruz2019memory} the authors proposed a memory-based reinforcement learning (MXRL) approach. The DRL agent is using episodic memory to make explanations using the probability of success and the transactions needed to reach the goal. The questions for the agent to answer were in the "why?" and "why not?" format. Each state-action pair is recorded on a list where the number of actions needed to reach the goal and the probability of success is calculated for the action by using the number of steps to success and total time steps. Using a simple grid game where the agent had to move to a goal position. {\color{black}The team found that this scheme generated explanations that non-expert users could understand. They recognised that this was only a limited situation and more work was needed to apply this to a more complicated problem.}

{\color{black}The next paper uses an estimation of Shapley values to assign a weighting to an input depending on how it affects the output. Heuillet et al. (2022)\cite{heuillet2022collective} calculated an approximation of the Shapley values using a Monte Carlo algorithm. They postulated that Shapley values could be a pertinent way to evaluate the contribution of different players in a multi-agent RL context. They proposed three research questions:}
\begin{itemize}
    \item Can Shapley value be used to determine how much each agent contributes to the global reward?
    \item Does the proposed Monte Carlo based algorithm empirically offer a good approximation of Shapley Values?
    \item What is the best method to replace an agent missing from the coalition (e.g., a random action, an action chosen randomly from another player or the "no operation" action)?
\end{itemize}

To answer these research questions, they used two games where different agents had to cooperate to complete the task. The first game was a "predator/prey" environment where the agents acted as the predators. The second game was a social game where the agents had to work together to produce as many apples as possible. They compared the Shapley values across up to six agents, where these agents had to cooperate to generate the largest global reward possible. To decipher which agents were producing the highest local rewards, they could analyse the Shapley values. 

The researchers found that the Shapley values were useful for explaining the multi-agent models that they were studying. They also found that the Monte Carlo derived approximations could be used as a suitable substitute for properly derived Shapley values. {\color{black}They believed felt that these explanations were useful to both researchers and developers. Additionally, they proposed that these explanations could also be presented to the general public as the intrinsic value of an agent. }The team also felt that these values could also be used as a method to detect bias in the DRL model when training. This would allow the analysis of the individual agents and an analysis of how they are completing the tasks. As for their final research question, they found that using a no-operation action was the most neutral and interaction-free method for this possibility. Using a random action, it was probable that a high negative reward could be achieved.

\begin{table}
    \centering
    \rowcolors{2}{gray!40}{gray!20}
    {\color{black}\begin{tabular}{p{5cm}p{1.5cm}p{1cm}p{1.5cm}p{4cm}}
        \hline
        Title of Paper & Authors & DRL Agent & XAI Method & Result of Study\\
        \hline
        Visualization of Deep Reinforcement Learning using Grad-CAM: How AI Plays Atari Games? & Joo et al. & A3C & Grad-CAM & Visualising the areas of attention while playing Atari video games. \\
        Towers of Saliency: A Reinforcement Learning Visualization Using Immersive Environments & Douglas et al. & N/A & Saliency Map & Creating a 3D saliency map for use with VR in Pommerman. \\
        Machine versus Human Attention in Deep Reinforcement Learning Tasks & Guo et al. & PPO & Saliency Map & Comparison between human and DRL agents attention and agent success. \\
        Explaining Reinforcement Learning to Mere Mortals: An Empirical Study & Anderson et al. & Model-free QL & Saliency Map & Using XAI to explain DRL agents actions in a RTS game to non-experts. \\
        Learning Sparse Evidence-Driven Interpretation to Understand Deep Reinforcement Learning Agents & Dao et al. & PPO & CAM/ SBRL & Using CAM and a Sparse Interpretable Module to explain DRL learning. \\
        Enhancing Explainability of Deep Reinforcement Learning through Selective Layer-Wise Relevance Propagation & Huber et al. & DQN & Saliency Map & Using selective LRP to only highlight the important saliency features. \\
        Towards Explainable Reinforcement Learning Using Scoring Mechanism Augmented Agents & Liu et al. & DQN/ DDQN/ A2C/ PPO & ARS & Using ARS they use score maps give to explanations as well as focus the DRL on the important areas during policy learning. \\
        Brittle AI, Causal Confusion, and Bad Mental Models: Challenges and Successes in the XAI Program & Druce et al. & DQN & CAMEL & Produced a UI to explain moves to human co-op players in StarCraft 2. \\
        Explainable Artificial Intelligence (XAI) for Increasing User Trust in Deep Reinforcement Learning Driven Autonomous Systems & Druce et al. & DQN & VEE/ DNTS & A small group were tested for the UI generated by the VEE and DNTS metrics to measure any increases in understanding. \\
        SDRL: Interpretable and Data-Efficient Deep Reinforcement Learning Leveraging Symbolic Planning & Lyu et al. & SDRL & Symbolic Planning & Marrying symbolic planning with DRL to improve interpretability. \\
        EDGE: Explaining Deep Reinforcement Learning Policies & Guo et al. & DQN & EDGE & Creating EDGE a module to create a self-explainable model by leveraging a Gaussian Process. \\
        Discovering symbolic policies with deep reinforcement learning & Landajuela et al. & Various & Symbolic Policy & Using RNN-based policy generator symbolic policies are generated that outperform DRLs while being more interpretable. \\
        Explaining Deep Reinforcement Learning Agents in the Atari Domain through a Surrogate Model & Sieusahai et al. & DQN & Surrogate Models & This study used sprites generated from pixel values to create a surrogate model. \\
        Moody Learners - Explaining Competitive Behaviour of Reinforcement Learning Agents & Barros et al. & DQN/ PPO & Moody & Analysing the Q-value selection and representing it on a Pleasure/Arousal scale to increase interpretability. \\
        \hline
    \end{tabular}}
    {\color{black}\caption{Summary of papers covered in Video Game Simulation, part one.}}
    \label{tab:video_table1}
\end{table}
\begin{table}
    \centering
    \rowcolors{2}{gray!40}{gray!20}
    {\color{black}\begin{tabular}{p{4cm}p{1.5cm}p{1.2cm}p{1.3cm}p{5cm}}
    \hline
    Title of Paper & Authors & DRL agent & XAI Method & Results of Study \\
    \hline
        Applying and Verifying an Explainability Method Based on Policy Graphs in the Context of Reinforcement Learning & Climent et al. & DQN & PGR & Applied PG to a DRL controlled Cartpole to produce explanations. \\
        A Novel Approach to Curiosity and Explainable Reinforcement Learning via Interpretable Sub-Goals & van Rossum et al. & MDP & Bicycle- GAN & Using a GAN to create a sub-goal focused agent that enables a higher level of explainability. \\
        CDT: Cascading Decision Trees for Explainable Reinforcement Learning & Ding et al. & PPO & Decision Tree & A CDT was created by combining feature learning and decision making DTs to achieve comparable performance to a DRL but with intrinsic interpretability. \\
        MoET: Mixture of Expert Trees and its Application to Verifiable Reinforcement Learning & Vasic et al. & DQN/ DDQN & Decision Tree & Introduced Mo{\"E}T a model based on mixing Mixture of Experts with DT to produce DRL like results. \\
        Learning Tree Interpretation from Object Representation for Deep Reinforcement Learning & Liu et al. & MDP & Decision Tree & Using a MCRTS algorithm to produce an IB-optimal mimic tree with DRL-like performance. \\
        Towards Interpretable-AI Policies Induction using evolutionary Nonlinear Decision Trees for Discrete Action Systems & Dhebar et al. & DNN & Decision Tree & Created a NLDT that performed on par with a DRL agent. \\
        LRP-based network pruning and policy distillation of robust and non-robust DRL agents for embedded systems & Luan et al. & DQN & LPR & Using relevancy scores generated by LPR a model is pruned to work more efficiently at playing Atari video games. \\
        Toward Interpretable Deep Reinforcement Learning with Linear Model U-Trees & Liu et al. & DQN & Decision Tree & Introduced a novel on-line LMUT learning algorithm producing a DT with DRL performance.  \\
        Explainable Reinforcement Learning via Model Transforms & Finkelstein et al. & DQN/ SARSA/ CEM & MDP Transforms & By analysing the MDP transforms explanations can be found for the DRL behaviour. \\
        DRLViz: Understanding Decisions and Memory in Deep Reinforcement Learning & Jaunet et al. & A2C & Memory Analysis & Produced a tool that allowed the analysis of memory to produce explanations for the DRL agent in DOOM. \\
        Accelerating the Learning of TAMER with Counterfactual Explanations & Karalus et al. & DQN & Counter- factuals & Showed that counterfactuals in TAMER improved training times when using human in loop training. \\
        Memory-Based Explainable Reinforcement Learning & Cruz et al. & SARSA & Memory based & Using episodic memory explanations were generated based on the probability of success and the number of transactions to the goal. \\
        Collective eXplainable AI: Explaining Cooperative Strategies and Agent Contribution in Multiagent Reinforcement Learning with Shapley Values & Heuillet et al. & DDPG/ MADDPG/ A3C & Shapley values & A Monte Carlo estimation of the Shapley values was used in collaborative games to produce explanations. \\
        \hline
    \end{tabular}
    \caption{Summary of papers covered in Video Game Simulation, part two.}}
    \label{tab:video_table2}
\end{table}

{\color{black}
From the summary of the studies of the first section presented in Tables 2 \& 3, most of the papers can be grouped into three major XAI fields: First is the use of saliency maps to show areas of attention of the DRL agents. The second is to create DTs that can mimic the performance of DRL agents while remaining more interpretable. The third is a mix of using mathematical functions to produce explanations and creating policies that can also serve as explanations.

The popularity of saliency maps is understandable as they are excellent at providing explanations for visual inputs. The video games that were explained using saliency mapping techniques had simple rules, and all information was provided by the input image. Using saliency mapping with a video game with more complex rules or off-screen information may not supply the same quality of explanations as those seen in the papers using simple Atari games. 

The use of DTs can be explained by the way the video games have a stringent set of easy-to-follow rules, which keep the complexity of the created DTs down. Once a decision tree becomes too complex, it loses its interpretability. Further research on whether DTs can be used in more complicated situations, either by increasing the complexity of the video game or by applying them to real world situations, would be useful.

The last section is devoted to the various methods that are grouped together. There is no trend in this section as the methods were quite varied. Heulliet et al\cite{heuillet2022collective} produced their own method of deriving Shapley values rather than the widely used SHAP package. There were no papers that used popular explainability methods such as SHAP or LIME. These more computationally expensive methods have been shown to have good results in different DRL applications. It is reasonable to suggest that these more computationally expensive methods may not be necessary, given that saliency mapping has been shown to perform well and at a much lower computational cost.}

\subsection{Vehicle Guidance}
The next area of study involves DRL networks and explainability in the guidance of vehicles. The vehicles covered in this section are; drones, wheeled robots, automobiles, and ships. The use of DRL networks in these tasks has a huge benefit as they operate in a continuous action space in which these types of NNs are specialised. XAI is imperative in this field as autonomous vehicles are likely to be the first public-facing systems and need high safety standards. Convincing both the general public and regulators of their safety will be vital to their adoption. This section will look at the four types of vehicles used in guidance, starting with the UAVs. 

The first paper to be looked at is by He et al (2021)\cite{he2021explainable}. In this work, they set up a UAV to autonomously navigate around a simulated environment and then used SHAP to generate contribution values for each of the kinematic UAV states or CNN layer activation maps. Using the SHAP values, they generated textual responses that justify the DRL agent's response to the goal and objects. By analysing these SHAP values, the section of the network that the highest impact on the DRL's decision could be identified. For example, an obstacle seen by the drone would activate the CNN layers producing a text response explaining the action as been effected by the highlighted CNN layers. The experiment reveals the importance of each section of the network to the agent's actions. The study also did a real-world test where it was confirmed to work when flying a UAV. Overall the SHAP CNN explainer was successful at describing the actions of the drone in both simulation and real-world testing. The authors found that this scheme for generating explanations produced good explanations that would be understandable to novice users. 

Next up is a paper by Guo et al.(2020)\cite{guo2020partially}, where they investigated how to use UAVs to provide coverage for 5G networks while maintaining efficiency in energy usage and signal optimisation. The model is rendered interpretable by the extraction of features in the hidden layers, this however only provides a small layer of explainability. They proposed a DDDQN with a Prioritised Experience Replay (PER) and a fixed Q-target that allows the model to maintain stability and prevent over-fitting while also offering performance gains. The team was able to show that this scheme was able to have more efficient UAV usage than typical UAV autonomous flight. They were also able to use the interpretability from examining the hidden layers to find the optimal drone deployment. The authors suggested that the next step was to increase the explainability of the model to understand the propagation of the features in the DDDQN.

Wang et al.(2020)\cite{chang2020coactive} look into how autonomous UAVs interact with human operators and large flocks of UAVs. They proposed an agent-based task planner that decomposes a task into a series of interpretable sub-goals. The team designed a simulation where drones had to avoid radar hot spots and with differing levels of human cooperation. They found that the swarm could complete complex tasks using the designed scheme. {\color{black}The explainability was improved by showing the decomposed tasks on a timeline, which helped to show how actions evolved over time, resulting in better interpretability.}

Stefik et al.(2021)\cite{stefik2021explaining} describe the COmmon Ground Learning and Explanation (COGLE) system in their paper. COGLE is an XAI method for a DRL system that delivers supplies to field units in mountainous areas. The explanations come in the form of "What", "Why" and "Where" questions. The "why" question depends on counterfactuals. The "what" question depends upon the mission profile that the UAV has chosen. Finally, the question of "where" is answered by analysing the map to highlight risks. The answers to these questions come from a visualisation of the map with risks highlighted and a section that gives narrative answers on why a route has been chosen. Overall this method produced good explanations of why the UAV had chosen a particular action. It was especially beneficial to have pre-decision explanations so the operator could make changes before any actions had taken place.

{\color{black}The next paper looks at a new explainable agent that is proposed by Parra-Ullauri et al.(2022)\cite{parra2022event}. They proposed Event-driven TEmporal MOdels for eXplanations (ETeMoX). ETeMoX is an architecture based upon Temporal Models (TMs) that look at the reasoning that the model makes through time and extract explanations that are history aware when asked. They use Complex Event Processing (CEP) and TMs to generate the explanations and present them to the end-user through plots and graphs. Using ETeMoX in three case studies that each leveraged a different type of DRL algorithm to control a UAV acting as Airborne Base Stations. They found that their methods created explanations at an acceptable level. They were explanations that both tracked the evolution of a metric and the relationships between them. They were also able to look at time windows to see how interest changed through that period. For future work, the team proposed to look at how the system could be made more efficient in computation. They also suggested looking at using these explanations to improve human-in-the-loop situations.}

Robotic GNC is the next area to be looked at. Nie et al.(2019)\cite{nie2019visualizing} look at visualising the behaviour of a swarm robot system. A swarm robotic system is a multi-robot system where many homogeneous autonomous robots act in tandem. They suggested using a Deconvolutional Network (Deconvnet) and Grad-CAM to visualise the decision-making process. The experiment was for the robots to visit two opposing places as much as possible in an allotted time. The team found that the two methods could interpret the policies learned by the DRL agent. They did point out that with actions coming from a DRL agent, it is difficult to say whether it is wrong or correct as there is no absolute standard. 

Roth et al.(2021)\cite{roth2021xai} use DT to give a layer of interpretability to the small robot they are guiding with a DRL agent. They go one step further with the transformation of their agent into a decision tree as they also apply performance improvements over what the DRL agent can achieve. Some of these improvements are smoothing out oscillations and frequency of immobilisation. They manage to do this without retraining the model. They demonstrated in their paper that they could provide interpretability and improve their agent with this technique.

Next, the paper by Josef et al.(2020)\cite{josef2020deep} looks at the guidance of a robot through an unknown obstacle course using a DRL agent while incorporating a self-attention module that provides explainability to the agent's actions. This paper mostly focuses on the new methods they implemented for designing a DRL agent for a robot. The use of self-attention for explainability was not a focus. It did find that the agent's attention was on the closest edge of a hazard to the robot, which is the expected response from this feature.

Xu et al.(2021)\cite{xu2021interpretable} use a symbolic DRL framework to improve data efficiency and interpretability when guiding a robot around an area. The framework consists of a high-level agent, a sub-task solver and a symbolic transition model. The high-level policy generates goals that the agent can complete to achieve the overarching goal. By separating the high-level and low-level goals, the high-level policy can focus on longer-term goals, and thus reducing sample complexity. The symbolic transition model allows interpretability because the logic rules for the model are readable by the operator. The sub-task solver is just the DRL that interacts with the environment to solve the short term goal that the high-level policy. The authors found that this method provides high efficiency and an interpretable model.

The following papers will be looking into automotive guidance, an important area of research with the first attempts at commercialised self-driving cars starting to appear. The first of these papers to be reviewed is by Soares et al.(2020)\cite{soares2020explaining}. In this paper, the authors attempt to provide explanations for their DRL model through rule-based approximation and visualisation. The rule-based approximations will be provided by a set of IF...THEN rules that will form an alternative interpretable model. A visualisation is provided to enhance the explanations. The experiments showed that this method was accurate and computationally efficient while allowing for good interpretability and thus validating the DRL agent.

As well as piloting a car, DNNs will also have to predict when accidents might occur and avoid them. Bao et al.(2021)\cite{bao2021drive} try to use a DRL agent to achieve this and provide visual explanations. The team's DRIVE model uses both a top-down and bottom-up visual attention mechanism to make observations from the dash-cam footage. It uses this information to produce an explanation from the attentive areas for the decision taken by the agent. The experiments showed that the DRIVE model generates state-of-the-art performance in real-world situations while providing good explainability.

Gajcin et al.(2022)\cite{gajcin2022contrastive} proposed using contrastive explanations to help decide which of the two DRL agents in the current situation to use. The best strategy can be hard to choose for a complex task when presented with options. These authors lay out a strategy for picking the better solution using contrastive explanations. They found they could find the preferences of the two models by comparing them. In their experiment, the researchers could see that the safer of the two models preferred to leave a more pronounced gap and travel at a slower speed. In future research, they wanted to move on to comparing more than two models at a time.

The paper by Liessner et al.(2021)\cite{liessner2021explainable} uses SHAP to describe the control of a car's speed. Shapley values have their origins in 1950s game theory which described how participants in a game could maximise their value. This concept was used to produce SHAP to generate explanations for NNs. The car in this experiment is driving down a one-lane highway and has to follow the speed limit changes that it comes across. Unsurprisingly the SHAP values showed that the current speed limit and the velocity had the highest impact on the model's choices. This method does provide great understanding through visuals for this simple problem and is promising for more complex problems.

{\color{black}In this paper by Wang et al.(2022)\cite{9922404} they look at using DRL for congestion avoidance. While the paper mostly deals with the method of their traffic routing they have also instilled a factor of explainability into their XRouting model. The car was given a grid of roads that had a distance, average speed, vehicle density, average travel time, and position encoding which is a measure of the distance to the end goal. From these attributes the XRouting model found the optimal route, this improved upon a PPO by a couple of percentage points in travel time, CO\textsubscript{2} emissions, fuel use, and route length. This was an improvement on the Dijkstra method by around 4\%. For explainability, they used a Gated Transformer (GTr) to rank the importance of the factors for each route. They were able to find that speed was the most important attribute to the XRouting model followed by vehicle density, travelling time, and end distance.}

The following paper looks at three methods for explainability in the DRL agent that controls the kinematics of a car. N.B Carbone (2020)\cite{carbone2020explainable} uses three explainability techniques to analyse their agent, SHAP, LIME, and Linear Model Trees (LMTs). LIME creates locally interpretable explanations for a single data point. By making these perturbations in the dataset, the importance of a feature can be deduced. Shapley values measure how much an individual feature in the NN affects the output. Several SHAP methods approximate the values instead of direct calculation. The calculation of the exact values is computationally expensive. This paper uses the Kernel SHAP, which builds on the LIME framework. LMTs produce interpretable results by creating a DT model approximating the targeted agent. This DT is more interpretable as the nodes follow interpretable rules. Much like the other methods that have used LMTs to approximate the models, they also found good performance. However, they found that there was a trade-off between the accuracy of the LMT and the interpretability as a more complex tree will be more accurate it would lose its interpretability. They found that the use of SHAP helped analyse the behaviour of the model, but it didn't produce the global policy insights into the agent that the LMT provided. The author then concludes that LMTs may become the gold standard in creating explanations for DRL agents.

There is only one paper {\color{black} within the reviewed time frame that investigates the use of explainable DRL agents for controlling watercraft,  this is} by L\o ver et al.(2021)\cite{lover2021explainable}. Again in this paper, the same three methods as the last paper are used: LIME, SHAP, and LMTs. These three methods appear as the benchmark methods for applying explainability to DRL models in the field. The experiment was to dock a ship to a quayside using the DRL agent and then measure the performance of the three explainers. The experiments showed that, for a task that needs real-time explanations, the LMTs achieved better solution. The two other methods were too computationally expensive. This meant the explanations were slow to be generated. They summarised that SHAP was better than LIME for post-hoc explanations though it may be susceptible to biased predictions.

\begin{table}
    \centering
    \rowcolors{2}{gray!40}{gray!20}
    {\color{black}\begin{tabular}{p{3.9cm}p{1.3cm}p{1.2cm}p{1.6cm}p{5cm}}
        \hline
        Title of Paper & Authors & DRL agent & XAI Method & Results of Study \\
        \hline
        Explainable Deep Reinforcement Learning for UAV autonomous path planning & He et al. & TD3 & SHAP & Using SHAP values to create attention maps for a simulator drone and then verified in real world testing. \\
        Partially Explainable Big Data Driven Deep Reinforcement Learning for Green 5G UAV & Guo et al. & D3QN & Feature Extraction & Using feature extraction behaviours of a drone swarm providing Wi-Fi hot spots became clearer. \\
        Coactive design of explainable agent-based task planning and deep reinforcement learning for human-UAVs teamwork & Wang et al. & CACER/ D3QN & Sub- Goals & In cooperation with humans a task planner split the decisions into sub-goals that could be shown on a timeline for greater interpretability. \\
        Explaining autonomous drones: An XAI journey & Stefik et al. & A2C & COGLE & Provided text and visual explanations on the choice of the route chosen for a UAV when factoring risk and efficiency. \\
        Event-driven temporal models for explanations - ETeMoX: explaining reinforcement learning & Parra-Ullauri et al. & DQN/ SARSA/ Q-learning & ETeMoX & Using their ETeMoX system they generated explanations through CEPs and TMs that produced plots and graphs for the user. \\
        Visualizing Deep Q-Learning to Understanding Behavior of Swarm Robotic System & Nie et al. & DQN & Grad- CAM/ Deconvnet & Using autonomous robots for a round trip task they used Grad-CAM and Deconvnet to provide explanations. \\
        XAI-N: Sensor-based Robot Navigation using Expert Policies and Decision Trees & Roth et al. & PPO & DT & Using DT the study uses this to modify the policies of the DRL without retraining. \\
        Deep Reinforcement Learning for Safe Local Planning of a Ground Vehicle in Unknown Rough Terrain & Josef et al. & Rainbow & Self attention module & The self attention module gave explanations but this was not the focus of the paper. \\
        Interpretable Model-based Hierarchical Reinforcement Learning using Inductive Logic Programming & Xu et al. & SMDP & Symbolic Transition & In this robotic guidance task the model was split into a high level and low level by implementing a symbolic transition model that increased explainability. \\
        Explaining Deep Learning Models Through Rule-Based Approximation and Visualization & Soares et al. & DDQN & IF THEN Rules & Guidance of 2D simulation of a car in traffic is modelled using IF...THEN statements to increase interpretability. \\
        DRIVE: Deep Reinforced Accident Anticipation with Visual Explanation & Bao et al. & SAC & Saliency Mapping & Using a DRL to anticipate car accidents saliency mapping monitors the attention of the model. \\
        Contrastive Explanations for Comparing Preferences of Reinforcement Learning Agents & Gajcin et al. & - & Contrastive Explanations & Two DRL agents were tasked with traffic navigation and explanations were generated by the different decisions by the models. \\
        Explainable Reinforcement Learning for Longitudinal Control & Liessner et al. & DDPG & SHAP & Using a DRL operated car tasked to keep a variable speed SHAP showed the important metrics. \\
        \hline
    \end{tabular}}
    {\color{black}\caption{Summary of papers covered in Vehicle Guidance. Part 1}}
    \label{tab:vehicle_table1}
\end{table}
\begin{table}
    \centering
    \rowcolors{2}{gray!40}{gray!20}
    {\color{black}\begin{tabular}{p{4cm}p{1.3cm}p{1.2cm}p{1.5cm}p{5cm}}
        \hline
        Title of Paper & Authors & DRL agent & XAI Method & Results of Study \\
        \hline
        XRouting: Explainable Vehicle Rerouting for Urban Road Congestion Avoidance using Deep Reinforcement Learning & Wang et al. & XRouting & GTr & A ranking of attributes was created and visualised to explain XRouting's decisions. \\
        Explainable AI for path following with Model Trees & Carbone & DDPG & Decision Tree/ SHAP & Using car guidance LMT were shown to have similar performance to SHAP for less cost. \\
        Explainable AI methods on a deep reinforcement learning agent for automatic docking & L\o ver et al. & PPO & LIME/ SHAP/ Decision Tree & 3 XAI methods were compared in a ship docking task. LMT was useful in real time while SHAP provided better explanations. \\
        \hline
    \end{tabular}}
    {\color{black}\caption{Summary of papers covered in Vehicle Guidance. Part 2}}
    \label{tab:vehicle_table2}
\end{table}
{\color{black}The summary of the papers covered in the vehicle guidance is shown in Tables 4 and 5. In this set of papers, there is a greater emphasis on more complex DRL agents compared to those used in the video games section. This might point to the more diverse environments encountered when operating vehicles in dynamic settings. These papers also used methods like LIME and SHAP when they were absent in the video game section. These methods improve upon the saliency methods used in the previous section as they can give explanations for non-visual inputs such as velocity or heading. These methods were also compared against LMTs in a couple of studies, these found that the LMTs were generally more useful in a real-time setting because of their lower computation cost when compared with LIME and SHAP. These papers did also find that some of the explanations were better with SHAP when computation cost was not a limiting factor. DTs were also well represented in this area of research thus underlining this method as one of the most important XAI techniques. Compared with the video game section the number of papers in this field was about a third of those in the previous section, this might suggest that the field of XAI is still in its early days as researchers test methods in more simple environments.}

\subsection{System Control}
The usefulness of DRL agents controlling time-continuous systems has been realised in the last few years. The ability of DRL agents to react quickly and effectively is impressive. This speed is needed when controlling processes that can happen quickly. These systems are most often safety-critical. Trust in the decisions made by these systems is vital. For this reason, explainability is a salient area of current research.

The first three papers will be looking at the application of DRL agents in the control of traffic light systems. The conventional method of controlling traffic lights based on a fixed timer can lead to inefficiencies if the level or pattern of traffic changes dramatically throughout the day. A system that can respond to changing circumstances is crucial. The first paper on this subject is by Schreiber et al.(2021)\cite{schreibertowards}. In their study, they use the SHAP framework to explain the policy of a traffic light control system. They measured their solution against a fixed time cycle of the traffic lights. The comparisons were based on average speed score, average wait time and average travel time. They found that the DRL model could improve the traffic flow by about 25\%. {\color{black}They also found that SHAP values described how the length of the queue of cars affected the model's decision-making regarding the length of delay for traffic lights. This allowed the researchers to see that the traffic light system was making reasonable decisions.}

Next, traffic signal control by Rizzo et al.(2019)\cite{rizzo2019reinforcement} also used the SHAP framework to help explain their DRL model. In this paper, the junction chosen was more complex when compared to the previous study. This junction was a four-way intersection with a roundabout and an underpass for one of the roads. These junctions are complex due to the many variables affecting traffic speed, such as any traffic that backs onto the main road from the slip road, slowing traffic that doesn't have to stop at the junction. The cumulative wait times using the DRL controlled lights were reduced from 3262s for the fixed short phase and 2594s for the fixed long phase down to 728s. A significant improvement over the timed lights. For the explainability, the SHAP values gave a detailed explanation of the decision to change the colour of the lights. The researchers could tell that the model did not place much importance on the traffic sensors from directions with low traffic flow and placed high importance on sensors with high traffic flow.

The last paper on traffic light control uses a different system, which is knowledge compilation. Wollenstein-Betech et al.(2020)\cite{wollenstein2020explainability} propose using knowledge compilation, which is a technique to build a Directed Acyclic Graph (DAG). {\color{black}The DAG is a representation of the logical theory of the model. The knowledge compilation assembles the unorganized logical theory of the DRL into a structured one increasing the explainability.} This DAG will use the deterministic Decomposable Negation Normal Form (d-DNNF). This method is computationally expensive but this analysis only has to be run once to obtain the explainable model. For the experiment, they used a four-lane highway intersection with the simulation time of 1.5 hours. They found that the d-DNNF DAG allowed them to get the likelihood of a state change in the system. They posited that using this tool in debugging a black-box system would let them check that the controller is logically sound.

In Zhang et al.(2022)\cite{9506997}, they look to use a DRL agent to control a power system's emergency procedures and explain the actions of this system using SHAP. They use the Deep-SHAP algorithm built upon the DeepLIFT algorithm, which takes a sample of states to approximate the Shapley values. They trained their DRL on a power system simulation. Using the SHAP values to make visualisations they found an increased understanding of which variables drove the state changes. They did find that raw SHAP values are hard to understand, but changing them into probabilities made them more understandable.

Next, the paper by Nunes et al.(2021)\cite{nuneshuman} described a system to help manage air traffic using Solution Space Diagrams (SSD). This system alerts the air traffic controller to any aircraft coming into proximity with each other and suggests an exit vector to the aircraft to avoid any collisions. Using this SSD as an input, the DRL agent outputs the required exit vector. This paper didn't talk about the explainability aspects of this technology. The visualisations used for the SSD did explain the DRL's actions without it explaining the DRL directly.

{\color{black}The next paper by Theumer et al.(2022)\cite{theumer2022explainable} looks at a production control system. The production system is built of many different agents that control the different aspects of the production system. To generate the explanations SHAP is used to evaluate the importance of the inputs to the models. Of their developed MARL system they saw an increase in lead time reward, inventory reward, and total reward ranging from 4\% and 40\% over competing models. They also found that SHAP allowed for reasonable explanations of the behaviours of the models.}

The final paper for review in the Systems control section is Kotevska et al.(2020)\cite{kotevska2020methodology}. They suggest a method for introducing interpretability to a Heating, Ventilation and Air Conditioning (HVAC) control system. To provide the explanations they used the LIME algorithm. The DRL model improved the efficiency of the HVAC system. However, the team found that this increase in efficiency was not enough to build trust in the system, though their use of the LIME algorithm allowed them to see inside the decision making of the control system. The LIME values showed which environmental factors were important in changing the system.

\begin{table}[ht]
    \centering
    \rowcolors{2}{gray!40}{gray!20}
    {\color{black}\begin{tabular}{p{3.9cm}p{1.8cm}p{1.1cm}p{1.4cm}p{4.8cm}}
        \hline
        Title of Paper & Authors & DRL Agent & XAI Method & Results of Study \\
        \hline
        Towards Explainable Deep Reinforcement Learning for Traffic Signal Control & Schreiber et al. & DQN & SHAP & Optimising a traffic signal junction using a DRL, SHAP was used to provide metrics on the importance of the different traffic elements. \\
        Reinforcement Learning with Explainability for Traffic Signal Control & Rizzo et al. & Monte-Carlo PG & SHAP & A complicated highway junction is controlled by a DRL agent. This is explained by SHAP, showing which traffic flows are important to the junction. \\
        Explainability of Intelligent Transportation Systems using Knowledge Compilation: a Traffic Light Controller Case & Wikkenstein-Beetech et al. & DQN & d-DNNF & Using knowledge compilation and a DAG the unorganised logic of a DRL is reorganised in a structured manner. The controller became more interpretable. \\
        Explainable AI in Deep Reinforcement Learning Models for Power System Emergency Control & Zhang et al. & DQN & SHAP & This study used SHAP to give a value of importance to the different inputs of the power system explaining the actions the system undertook. \\
        Human-interpretable Input for Machine Learning in Tactical Air Traffic Control & Nunes et al. & DDQN & SSD & To improve interpretability in AI ATC SSDs are used to explain decisions. \\
        Explainable Deep Reinforcement Learning For Production Control & Theumer et al. & MARL & SHAP & In this production system SHAP was used to explain the action of a multi-agent system. \\
        Methodology for interpretable reinforcement learning model for HVAC energy control & Kotevska et al. & DQN & LIME & LIME was used to explain decisions taken by the HVAC system. \\
        \hline
    \end{tabular}
    \caption{Papers covered in System Control.}}
    \label{tab:system_table}
\end{table}

{\color{black} The papers in this section looked at controlling systems that only take inputs as values from various sensors. This means that no saliency mapping techniques show up in this section. Techniques like SHAP and LIME feature heavily, as they will give explanations for all the inputs to the DRL model by assigning numerical values to each one. The values generated from these XAI methods can be then used in a number of ways and visualised in the most appropriate manner for the style of explanations wanted. The success of explaining the traffic light control simulations shows a great example of the use of XAI being able to enhance DRL solutions by providing insights into decisions made. The explanations mostly agreed with human intuition which confirms the DRL is behaving in a reasonable manner. The use of XAI and DRL in air traffic control systems could be an area for future growth. The number of commercial passenger aircraft is still increasing and the introduction of autonomous air taxis and delivery drones will increase the workload for air traffic controllers. In these mission critical roles, trust is likely to be of utmost importance underlining the need for XAI methods. }

\subsection{Robotic Manipulators}
The next application to be looked at is robotic manipulators controlled by DRL agents. The use of these networks has grown in popularity recently, as the ability of these control systems allows the agent to adapt to situations that it has not necessarily seen before. {\color{black}These features would allow for more generalised robots in assembly lines that can complete various tasks instead of specialised robots.}

The first three papers in this section use the SHAP algorithm to describe their DRL agents. The first is by Remman et al.(2022)\cite{remman2022causal}. Their study looks at how the robotic lever manipulator system can be described using SHAP. Their application of SHAP is an alteration of the Kernal-SHAP, as that method only describes the direct effect that features have on the output. The team suggests creating a Causal SHAP that looks at the indirect effects that a feature has on the output. To accomplish this, Causal SHAP alters the sampling method used by Kernel SHAP by using a partial causal ordering that captures these indirect effects. The robotic manipulator used is the OpenMANIPULATOR-X, which they tasked with moving a lever from one position to another. They showed that Causal SHAP describes both the direct and indirect effects that a feature can have on the output. They also showed that by using Casual SHAP they can generate better explanations. They did recognise that these explanations are not well suited to non-expert users but are helpful for data scientists and researchers.

The second paper on robotic manipulators using SHAP is by Remman et al.(2021)\cite{remman2021robotic}. This paper has a similar experimental setup to the previous study but focuses on using a DDPG algorithm with a Hindsight Experience Replay. They use the regular SHAP implementation to test their DRL agent. They make some conclusions about using  SHAP values with this DRL agent. They found that some variables featured prominently in the SHAP values as expected, whereas others, namely the joint variables, did not. They theorise that the lack of assumed independence between states may lead to this, as there is a correlation between states. 

The final SHAP paper for this section is from Wang et al.(2020)\cite{wang2020attribution}. Their study applies the SHAP method to describe an automated crane system. The mass that the crane is lifting can swing, which complicates this situation. The DRL agent has to account for the momentum that this object will have and learn to be smooth as possible. This paper was too short for any grand conclusions, though they found that the SHAP values did generate the explanations as predicted.

Next in the paper, Iucci et al.(2021)\cite{iucci2021explainable} address XAI in Human-Robot Collaboration (HRC) scenarios. They do this by combining two methods. First, they use Reward Decomposition to break down the reward function to give insight into the factors that influenced the agent's decisions. The second is an Autonomous Policy Explanation (APE) that produces natural language responses to queries about the robot's behaviour. To differentiate between the rewards an action generated, it was classified into five sub-types. The APE uses a series of logic statements placed onto a grid to produce a natural language explanation. Using these approaches, the authors found that they could produce satisfactory explanations. These types of explanations were useful when debugging the DRL agent.

The following paper is by Schwaiger et al.(2021)\cite{schwaiger2021explainable}. This study investigated how to extract explainability from the DRL agent of a robotic manipulator performing a pick-and-place task. The aim was to find the dimensions of the robot arms from the DRL agent to open up the black box. The authors theorise that the DRL agent must find these dimensions to learn a task. If extracting the dimensions is possible, then generating explanations could be done. They found that they could accurately identify the lengths with the highest average error for a section being 0.25\%. They felt that this justified their hypothesis. 

The final paper in the robotic manipulator section is by Cruz et al.(2021)\cite{cruz2021explainable}. They propose using goal-driven explanations to add interpretability to their DRL agent. They suggest these goal-driven explanations to try to create explanations that would be understandable by a non-expert user. They look to be able to answer the questions of why? And why not? Using the probability of success (P$_{s}$) for an action, they generate an explanation of the favouring of one action over another. The three ways of producing this probability of success are the Memory, Learning, and Introspection-based approaches. 

The memory-based approach uses a list of state-action pairs that produce the P$_{s}$. The method has a downside of the size of the memory growing as the number of episodes grows. 

The learning-based approach uses the agent's learning process to generate the P$_{s}$. The algorithm learns as the agent does by using P-values instead of Q-values which are usually the output of a training process. This process does add some overhead in the memory and does increase computing time during the learning phase. 

The introspection-based approach tries to remove all memory overhead by calculating the P$_{s}$ using a numerical transform to the Q-value directly. Q-value is used to approximate the reward at that point and determine how far the value is from the total reward. This can be converted to P$_{s}$ by applying a transform to get the correct shape of the curve. 

Testing these methods was done using three robotic tasks, two simulations and one real-world test. They found that all three explainability methods produced similar results and thus validated their introspection method as a good alternative with a light computing load. They found that the explanations were not perfect. The authors suggest this could be an issue later where the user becomes too trusting in the explanations, or wrong explanations destroy all trust between the agent and the user. {\color{black}This could especially be a problem in scenarios where explanations are required for non-experts. }In the future, they want to look at reward decomposition to generate more reward signals to produce explanations. 

\begin{table}
    \centering
    \rowcolors{2}{gray!40}{gray!20}
    {\color{black}\begin{tabular}{p{4cm}p{1.3cm}p{1.2cm}p{1.5cm}p{5cm}}
        \hline
        Title of Paper & Authors & DRL Agent & XAI Method & Results of Study \\
        \hline
        Causal versus Marginal Shapley Values for Robotic Lever Manipulation Controlled using Deep Reinforcement Learning & Remman et al. & DDPG & SHAP & In this paper, the manipulators actions were explained by SHAP showing both direct and indirect effects upon decisions. \\
        Robotic Lever Manipulation using Hindsight Experience Replay and Shapley Additive Explanations & Remman et al. & DDPG & SHAP & This paper expands on the other by introducing a Hindsight Experience Replay to improve training and uses SHAP to provide explanations. \\
        Attribution-based Salience Method towards Interpretable Reinforcement & Wang et al. & DQN & SHAP & Using a crane model SHAP is used to explain the movement of the crane while keeping the load under control. \\
        Explainable Reinforcement Learning for Human-Robot Collaboration & Iucci et al. & DQN & Natural Language/ Reward Decomposition & In a Human Robot Collaboration the DRLs actions were described by breaking down the reward and producing natural language responses from Boolean statements. \\
        Explainable Artificial Intelligence For Robot Arm Control & Schwaiger et al. & Not specified & Dimension Extraction & In a pick-and-place task the dimensions of the robot arm were extracted from the model allowing it to become more explainable. \\
        Explainable robotic systems: Understanding goal-driven actions in a reinforcement learning scenario & Cruz et al. & SARSA & POS & Using a memory-based probability of success metric this paper generates explanations on a number of robot tasks. \\
        \hline
    \end{tabular}
    \caption{Papers covered in Robotic Manipulators.}}
    \label{tab:robot_table}
\end{table}

{\color{black}The use of DRLs in manipulators is probably one of the areas of DRL research that will be adopted by industry first. Having one manipulator that can do many tasks is more efficient than having many manipulators that can only do one highly constrained task. Again, SHAP was used heavily in this section, marking it as one of the top XAI methods currently available. The SHAP values gave indications to the researchers which inputs were more important to a manipulator in making its decision. Apart from the three papers that covered SHAP, the other papers looked at extracting the explanations directly from the models. The final method was particularly promising as, when used along with the training of the model, its interpretability was greatly improved. In the fifth paper in this section, the ability to gain the dimensions of the manipulator from the DRL model is a promising breakthrough that could be applied to many different DRL research fields.}

\subsection{Network Solutions}
The ubiquity of mobile networks has become part of modern life. With the introduction of 5G as the latest standard, this is becoming more prevalent. The need for more and more data bandwidth has focused researchers on efficiency gains. Therefore, the use of DRL agents to find these optimisations is proving to be a rich vein for researchers. Using DRL agents leads to the question of explainability, especially in equipment that has become so important in modern life. With the needs of regulators and end-users, this is an area that requires research. As this is quite a niche application, there are currently only two papers covered within the time frame of this research.

The first paper by Li et al.(2021)\cite{li2021explainable} aims to defend their network against threats by using explainable DRL agents. The study looks at the edge devices, which are most susceptible to attack, and builds an agent to respond to the threats and allocate resources correctly. The agent used to perform this defence is a DDQN. To generate the explanations for the DDQN, they used LIME. LIME uses perturbations to the features to see the impact on the output and generates a probability of an outcome. In this paper, they managed to produce a model that helped deal with attacks on a network and successfully used the LIME package to explain the actions that this model chose.

The second paper is by Vijay et al.(2019)\cite{vijay2019secured}. Here, they look to overcome the inherent security issues in a 5G network. The authors look to develop a Main Aggregator Server (MAS) with a DQN that aggregates the responses and makes final decision in network allocations. Though the paper suggests that this approach will improve the use of XAI in this field, there is no suggestion on how to implement XAI into this particular type of agent, and further research is needed.

\begin{table}
    \centering
    \rowcolors{2}{gray!40}{gray!20}
    {\color{black}\begin{tabular}{p{4cm}p{1.3cm}p{1.2cm}p{1.5cm}p{5cm}}
        \hline
        Title of Paper & Authors & DRL Agent & XAI Method & Results of Study \\
        \hline
        Explainable Intelligence-Driven Defense Mechanism against Advanced Persistent Threats: A Joint Edge Game and AI Approach & Li et al. & DDQN & LIME & Looking at protecting edge network devices with DRL they use LIME to generate explanations. \\
        Secured AI guided Architecture for D2D Systems of Massive MIMO deployed in 5G Networks & Vijay & DQN & AI Aggregator Unit & Proposing a network architecture that uses XAI to help run a 5G network controlled by a DRL agent. \\
        \hline
    \end{tabular}
    \caption{Papers covered in Network Solutions}}
    \label{tab:network_table}
\end{table}

{\color{black}The papers for this application are more limited so any trends in the field are impossible to decipher. The first of the two papers used a method of XAI that is familiar to many of the papers in this review in applying LIME. The other paper was more theoretical as the study only suggested a network architecture that could be applied in 5g networks. It is therefore difficult to examine their claims of such a system without it being constructed first.}

\subsection{Medical Applications}
 {\color{black}The only paper in this section does involve some use of Atari games for preliminary studies. The paper's main focus is on using DRL agents to diagnose HIV patients while maintaining a level of explainability to experts. There is only one paper focusing on applying XAI to medical sciences, as this has been an area of interest for AI.}

Mishra et al.(2022)\cite{mishra2022not} tackled XAI by designing a PolicyExplainer visualisation, using a decision tree classifier to create the visualisation. This decision tree was able to track the features that led to classification results. They proposed a text-based solution, so the software could answer the questions why? Why not? And when? The authors felt that this is the best way to explain the agent's behaviour to an operator unfamiliar with ML. Using the visualiser in three tasks, the first two were simple games. The third was a study into the prediction of HIV infection. They found that the PolicyExplainer gave good explanations in the first two tasks and then used experts in the HIV domain to use the software to judge its usefulness. The experts found that the software gave good explanations that made sense and were informative. The authors felt that the interface currently is limited by state and action scalability. The plan is for further research to solve this problem by looking at ways to select the most crucial state features.

\begin{table}
    \centering
    \rowcolors{2}{gray!40}{gray!20}
    {\color{black}\begin{tabular}{p{4cm}p{1.3cm}p{1.2cm}p{1.5cm}p{5cm}}
        \hline
        Title of Paper & Authors & DRL Agent & XAI Method & Results of Study \\
        \hline
        Why? Why not? When? Visual Explanations of Agent Behaviour in Reinforcement Learning & Mishra et al. & Model-Based RL/ DQN/ Q-learning & Policy Explainer & Using the PolicyExplainer across three different scenarios including a medical application answers were generated to why, why not, and when questions. \\
        \hline
    \end{tabular}
    \caption{Papers covered in the medical applications.}}
    \label{tab:other_table}
\end{table}

{\color{black}The PolicyExplainer proposed by the paper looks like a robust method for providing explanations and was shown to work in different fields. This paper is also notable for the inclusion of a medical DRL application. This is bound to be an important field in the future, so hopefully, more papers are coming in this field.}

\section{Limitations of the Current Methods}
From these papers covered in the review, there are some limitations.  First, the applications for some of these XAI methods are working on uncomplex models. These uncomplex models may not scale to more complicated tasks or may be of limited value by being too specific to one task. For example, an explanation that works in an Atari game might not hold up in a more complex game with more actions available and more consequences to those actions. This problem appears more in the DT based XAI methods as a more complex system produces a large DT that becomes more uninterpretable as the DT grows. Second, the numerical XAI methods SHAP and LIME can be used for more complex systems, though they have to deal with high compute loads limiting them to situations where the explanations are not time sensitive. 

Finally, there is the problem of judging the explanations provided by the schemes described here. Only a few of the 59 papers used some form of survey group to test the explanations gained, and few involve large enough samples to make of experts conclusions. Surveying experts is also a problem due to the small number of experts in any given field. With these trust and understanding tests, it is much easier to gain trust in a model performing a simple task that has no bearing on the person observing. To build the trust required for AI to be accepted by the general public by producing explanations with real impact on the individual, such as an autonomous car explaining decisions to passengers. However, with the models still being simple, it is harder to build a deeper level of trust that is impactful. 

\section{Future areas of study}
{\color{black}Though these papers put forward many excellent XAI methods in a range of fields, there are still some areas where DRL methods have not been explainable.} Future studies must apply to more complex systems to see if solutions are scalable and produce viable explanations. One area where it was surprising that there was a lack of research was medical science. Only one paper looked at this field when it seems like an area that could benefit from XAI. It is crucial to trust models directly linked to people's health. In the future, it would be good to see more papers covering this. {\color{black} There are also areas in DRL that have not utilised XAI methods in the period that the review covered that could benefit from the methods. These DRL applications are in research fields like audio processing and natural language detection and tone analysis.}

Furthermore, testing the explanations on the people they are designed for should be carried out, or making comparisons on which types of explanations certain groups prefer would be interesting. Studies that cover this would allow the people designing systems that interact with the outside world to understand what approaches work when choosing the appropriate XAI solution.

\section{Conclusion}
The applications of the different XAI methods in the different fields of study can now be summarised. In the video game examples, there were the widest range of different methods. {\color{black} The simplicity of the video games allows for XAI methods to be tested out before trying them on more complicated simulations or real-world applications. There is also a large body of work running XAI methods on these video games, so comparisons are easily made.} Among the XAI methods, the construction of symbolic networks in the form of DT was a popular method to mimic the DRL agents. The use of these interpretable networks is expected as video games can have simplistic state spaces and limited actions that can be approximated easily. Another method that was popular with these types of applications was using saliency methods. With the visual element in these video games, a purely visual method like saliency maps is expected to be a popular method. It allows a person looking for an explanation to look directly at where on the screen the DRL model finds most important during the decision. This is one of the better ways of providing a succinct explanation. 

In the field of vehicle GNC, the XAI methods used numerical methods such as SHAP values more than in video games and in these more complex spaces. There were fewer DTs used as the DRL models became more complicated and thus more difficult to mimic. There were still visual methods used on the images recovered from the sensors. These methods were sometimes mixed with other methods, like He\cite{he2021explainable}, they used GRAD-CAM and SHAP values to explain the drone's attention using Grad-CAM and then provide more complex explanations using the SHAP values.

For system control, the use of SHAP values seemed to be the most popular method. There are no visual inputs to these systems, so methods such as saliency maps have no value. There were no DTs either, so this could be an area to explore in future studies. The researchers felt that SHAP and LIME provided good explanations for this particular application of DRL agents. Measuring the importance of an input to the model's decisions fits nicely with these methods due to the inputs generally being numerical values and the explanations being for trained operators.

Robotic manipulators also heavily used SHAP values for explanations, this suggests that, particularly in applications without a visual element, SHAP values are a great tool to deliver explanations to DRL agents. Along with SHAP values, there were methods used that extracted the features of the robotic arm from the network to break the black box directly. One method managed to increase interpretability when training the DRL agent, which could be a new avenue to explore. 

{\color{black}In the medical field, it is hard to draw any conclusions about which XAI methods are best as there is only one example. The recommendation is for further research into the medical field by applying some of the XAI methods that have been described in other applications.}

As ML and deep reinforcement learning become prevalent, it is imperative to give explanations for the actions they take. The studies shown here show good progress in several fields and provide explanations that can be useful for experts and non-experts alike. DTs have been shown to work well in many different situations, though scalability needs addressing along with working with more complex DRL models. Using metrics such as SHAP and LIME has been shown to be effective in many different scenarios. For example, in applications with no visual input like traffic light systems, power management, and production controls. Finally, visualisation methods for generating saliency maps such as Grad-CAM have been shown to be very effective in the video games tested.

There is further work in proving the scalability of these ideas and proving that they can be understood well by their respective audiences. Supplying explanations for AI solutions must continue to improve. Improvements from better explanations, using better visualisation methods, and producing well-articulated explanations. With these explanations, the public can accept the role of ML and its adoption can begin.}
\bibliography{reviewref}
\bibliographystyle{ACM-Reference-Format}

\end{document}